\documentclass[runningheads]{llncs}

\usepackage[final]{eccv}

\usepackage{eccvabbrv}

\usepackage{graphicx}
\usepackage{booktabs}
\usepackage{multirow}
\usepackage{colortbl}
\usepackage{makecell}
\usepackage{array}
\usepackage{pifont}
\usepackage{tikz}
\usepackage{soul}
\usepackage{algorithm}
\usepackage{algpseudocode}
\usepackage[most]{tcolorbox}
\usepackage{enumitem}
\usepackage{wrapfig}
\let\llncssubparagraph\subparagraph
\let\subparagraph\relax
\usepackage{titlesec}
\let\subparagraph\llncssubparagraph
\titlespacing*{\section}{0pt}{2.5ex plus 0.8ex minus 0.3ex}{1.2ex plus 0.3ex}
\titlespacing*{\subsection}{0pt}{2.0ex plus 0.6ex minus 0.2ex}{0.8ex plus 0.3ex}
\titlespacing*{\subsubsection}{0pt}{1.5ex plus 0.4ex minus 0.2ex}{0.6ex plus 0.2ex}
\titlespacing*{\paragraph}{0pt}{1.2ex plus 0.3ex minus 0.1ex}{0.5em}

\setlist{leftmargin=*, topsep=3pt, itemsep=2pt}

\usepackage[expansion=false]{microtype}

\usepackage{sidecap}

\definecolor{headerblue}{RGB}{41, 98, 255}
\definecolor{lightgray}{RGB}{245, 245, 245}
\definecolor{darkgray}{RGB}{100, 100, 100}
\definecolor{bestgreen}{RGB}{0, 128, 0}
\definecolor{accentblue}{RGB}{230, 240, 255}
\definecolor{best}{HTML}{D4EFDF}
\definecolor{lightgreen}{RGB}{230,255,230}

\sethlcolor{green!10}

\newcommand*\circled[1]{%
  \textcircled{\raisebox{-0.5pt}{#1}}%
}

\newtcolorbox[auto counter, number within=section]{highlightbox}{
    colback=gray!8, colframe=black, sharp corners,
    left=2pt, right=2pt, top=2pt, bottom=2pt,
    enhanced, boxrule=0.5pt, width=\linewidth,
    sharp corners, drop shadow=black!50!white
}

\newtcolorbox[auto counter, number within=section]{theoremproofbox_blue}[2][]{
    colback=cyan!6, colframe=cyan!20, coltitle=black, coltext=black,
    enhanced, sharp corners,
    left=2pt, right=2pt, top=2pt, bottom=2pt,
    boxrule=0.5pt, width=\linewidth,
    drop shadow=black!50!white, fonttitle=\bfseries, title={#1},
}

\newtcolorbox{theorembox_blue}{
    colback=blue!5, colframe=blue!60, fonttitle=\bfseries,
    title=Theorem 1 (Existence and Uniqueness), sharp corners, boxrule=1pt
}

\newtcolorbox{theorembox_green}{
    colback=green!5, colframe=green!60, fonttitle=\bfseries,
    title=Theorem 2 (Closed-Form Solution), sharp corners, boxrule=1pt
}

\newtcolorbox{theorembox_red}{
    colback=red!5, colframe=red!60, fonttitle=\bfseries,
    title=Theorem 3 (Flow Trajectory Stability), sharp corners, boxrule=1pt
}

\newtcolorbox{warningbox}{
    colback=red!10, colframe=red!40, coltext=black,
    width=\linewidth, boxrule=0.5pt, enhanced, sharp corners,
    left=6pt, right=6pt, top=4pt, bottom=4pt,
    fontupper=\normalsize\bfseries,
}

\newtcolorbox[auto counter, number within=section]{theoremproofbox_red}[2][]{
    colback=red!6, colframe=red!25, coltitle=black, coltext=black,
    enhanced, sharp corners,
    left=2pt, right=2pt, top=2pt, bottom=2pt,
    boxrule=0.5pt, width=\linewidth,
    drop shadow=black!50!white, fonttitle=\bfseries, title={#1},
}

\newtcolorbox[auto counter, number within=section]{theoremproofbox_green}[2][]{
    colback=green!6, colframe=green!25, coltitle=black, coltext=black,
    enhanced, sharp corners,
    left=2pt, right=2pt, top=2pt, bottom=2pt,
    boxrule=0.5pt, width=\linewidth,
    drop shadow=black!50!white, fonttitle=\bfseries, title={#1},
}

\newtcolorbox{proofbox}{
    colback=gray!3, colframe=gray!20, boxrule=0.5pt, arc=2pt,
    left=4pt, right=4pt, top=4pt, bottom=4pt
}

\tcbset{
  promptbox/.style={
    colback=gray!5, colframe=black!60, coltext=black,
    width=\textwidth, boxrule=0.4pt, enhanced, sharp corners,
    left=8pt, right=8pt, top=8pt, bottom=8pt
  }
}

\usepackage[pagebackref,breaklinks,colorlinks,citecolor=eccvblue]{hyperref}

\usepackage[accsupp]{axessibility}

\begin{document}


\title{Introspective Attention Modulation for Safe Text-to-Image Generation}

\titlerunning{Introspective Attention Modulation for Safe T2I Generation}

\author{Basim Azam\inst{1} \and
Hossein Rahmani\inst{2} \and
Naveed Akhtar\inst{1}}

\authorrunning{B. Azam et al.}

\institute{
The University of Melbourne, Australia\\
\email{\{basim.azam, naveed.akhtar1\}@unimelb.edu.au}
\and
Lancaster University, United Kingdom\\
\email{h.rahmani@lancaster.ac.uk}}

\maketitle


\begin{figure}[t]
    \centering
    \includegraphics[width=0.95\textwidth]{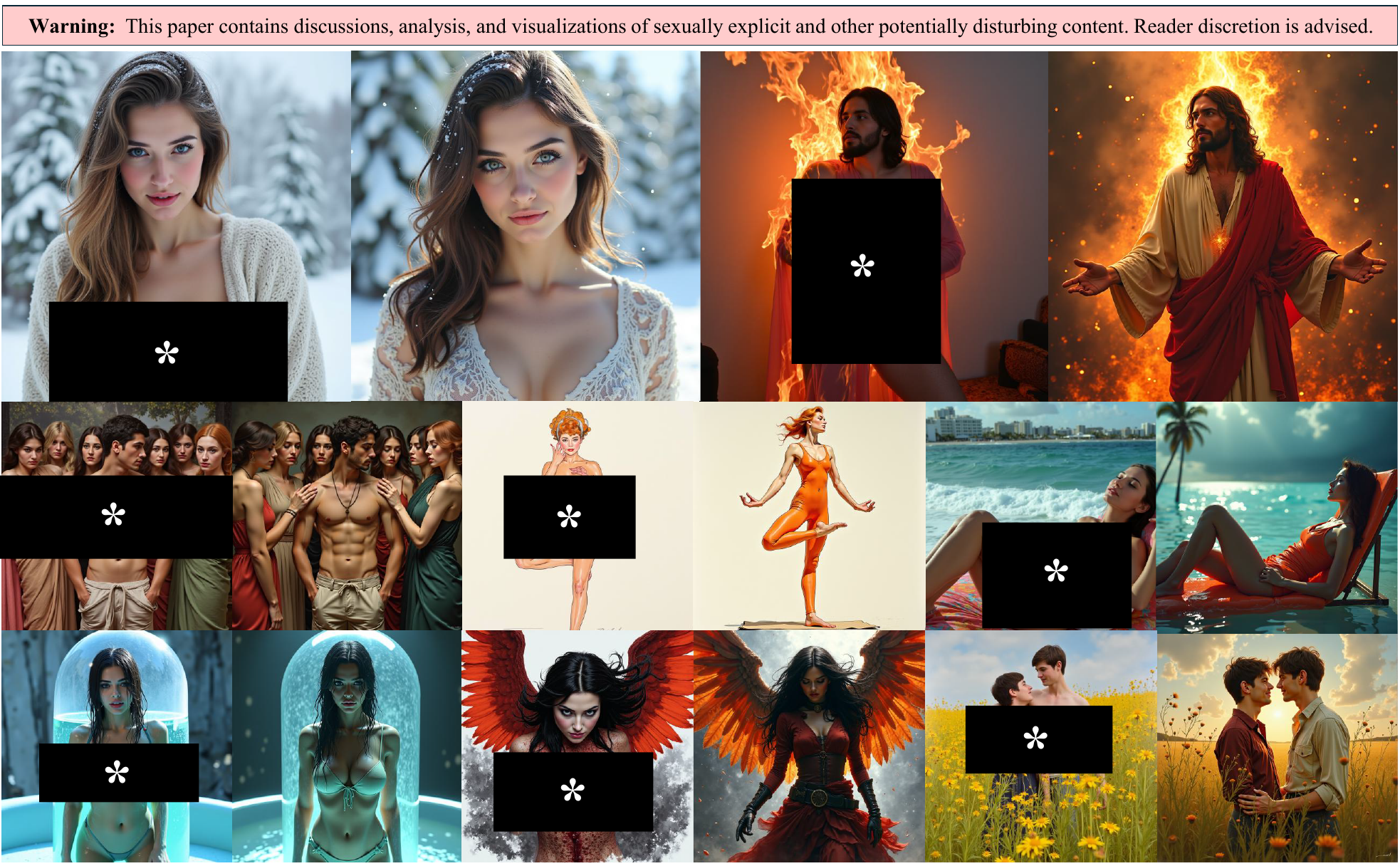}
    \vspace{-0.2em}
    \caption{
    Proposed method retains high image quality and semantic alignment while successfully suppressing unsafe content.
    Representative examples of flux.1-dev outputs are shown. For each image pair: \textit{{Left}} is baseline model output and \textit{{Right}} is the result with our intervention. {Blur is manually added by the authors for discretion.}}
    \label{fig:teaser}
    \vspace{-1.2em}
\end{figure}


\begin{abstract}
State-of-the-art flow based text-to-image (T2I) models exhibit remarkable generative abilities but remain vulnerable to producing unsafe  content. Prior safety efforts range from concept erasure and prompt filtering to classifier-based gating. However, simple techniques like parameter efficient adaptations of the models easily bypass  such guardrails.
We introduce a unique principled approach that achieves safety
by regulating the model's attention dynamics through  inference-time introspection,  exhibiting intrinsic    robustness. Our method
analyzes and rebalances attention activations throughout image synthesis, steering generations away from unsafe concepts while preserving semantic alignment. This introspective control ensures safety of deployed models.
Across standard and adversarial safety benchmarks, our approach achieves remarkable safety scores while maintaining or even improving alignment and perceptual quality. Our results reveal that attention-space regulation offers a considerably more promising path to safer diffusion transformer based image generation than the existing concept erasing mechanism.
{Our code can be accessed at \textcolor{magenta}{https://basim-azam.github.io/iam/}     
}
\end{abstract}

\vspace{-5mm}
\section{Introduction}
\vspace{-1mm}
Text-to-image (T2I) diffusion models ~\cite{Rombach2022,Ramesh2022,Saharia2022,song2021scorebased} have achieved impressive photorealistic image synthesis abilities. However, these models are often trained on uncurated web-scale data~\cite{Schuhmann2022}, which introduces serious safety vulnerabilities. Models generate violent, explicit, and sexual imagery even with seemingly innocuous prompts~\cite{rando2022redteaming,Yang2024SneakyPrompt}.
Whereas safety guardrails of recent models  are progressively improving \cite{esser2024flux},  parameter-efficient fine-tuning techniques like LoRA~\cite{hu2021lora} and DreamBooth~\cite{ruiz2023dreambooth} have allowed users to bypass the original safety  with minimal pre-deployment finetuning~\cite{zheng2024imma,yang2024guardt2i}. As advanced architectures like DiT transformers~\cite{peebles2023dit} and flow-matching  models~\cite{lipman2022flow,liu2022rectified,esser2024flux}  see wider deployment due to the realism of generated content, the tension between generation quality and content safety becomes increasingly critical.


Current safety mechanisms for T2I generation broadly fall into three  categories. \textit{External filtering} approaches rely on prompt blacklists and CLIP-based NSFW  classifiers~\cite{compvis2022safety,OpenAI2023SystemCard} to screen inputs or outputs. Yet, they prove brittle to adversarial prompting, 
including paraphrasing, obfuscation, and jailbreak attacks~\cite{Yang2024SneakyPrompt}. 
\textit{Training-based suppression methods} attempt to purge unsafe concepts from model's internals, either via targeted {concept erasure} ~\cite{gandikota2023erasing,gandikota2024unified,gao2024eraseanything, Azam_2025_CVPR, li2024selfdiscovering,tsai2024ringabell} or via safe 
fine-tuning of model weights~\cite{Schramowski2023}. Such approaches are computationally expensive and limited to the early diffusion variant they were developed for. 
Current \textit{inference-time methods} \cite{Schramowski2023, Liu2024} require retraining that also causes collateral semantic erasure. Moreover, weight-editing fixes can be undone by adapters that override the safety alignment~\cite{zheng2024imma,yang2024guardt2i}. 
In essence, these safety methods operate as post-hoc patches without a mechanistic understanding of  unsafe content emergence   during the generation, leaving significant gaps in reliability.  


We envisage that the diffusion models themselves hold the key to their  safety.
Recent advances in mechanistic interpretability for diffusion networks~\cite{Tang2023DAAM,park2024dissecting,conmy2023circuits,elhage2022superposition} 
affirm that the {attention mechanisms} of these models encode rich internal structure throughout the denoising process. Cross-attention maps, for example, show where the model attends to specific text tokens when constructing visual regions, while self-attention patterns enforce spatial and semantic coherence. Naturally, these meaningful structures must also hold meaningful cues for potentially unsafe concept formation in image synthesis.
This has led us to ask: \textit{can a diffusion model progressively introspect  its own attention and self-correct if it starts focusing on unsafe concepts?} 

We achieve an affirmative answer to this question  with our introspective attention modulation technique, 
which provides a principled  safety framework to intervene at the image synthesis process  itself. Without any retraining of the original model, our method enables a diffusion model to suppress unsafe content by regulating its attention. 
Our contributions are summarized as follows:

\begin{itemize}[leftmargin=*,itemsep=1pt]
    \item We introduce a first-of-its-kind attention based introspective safety mechanism for inference-time safety of high quality image generation with the modern diffusion transformer based T2I models. Our approach  offers a unique  principled guidance for the attention evolution during the image synthesis process. 
    \item To realize our method, we develop mechanisms of identifying and extracting unsafe attentions in diffusion transformer image synthesis process, and method to regulate them with a sparse logistic regressor. Moreover, we implement the optimization process and devise an introspection ratio control to provide a handle over safety-quality trade-off in our method. 
    \item We extensively evaluate the proposed  technique on the standard benchmarks  I2P~\cite{Schramowski2023}, SneakyPrompts~\cite{Yang2024SneakyPrompt}, MMA-Diffusion~\cite{yang2024mma} and UnlearnDiffAtk~\cite{zhang2025unlearnDiffAtk}, considerably improving safety and quality over the popular existing safety methods UCE~\cite{gandikota2024unified}, ESD~\cite{gandikota2023erasing}, EraseAnything~\cite{gao2024eraseanything}, FlowEdit~\cite{kulikov2025flowedit}, and \textcolor{black}{MCE~\cite{lu2025mce}}. 
    Our results establish for the first time that attention based regulation is significantly more effective for diffusion transformer based modern architectures than the conventional concept erasing. 
\end{itemize}

\section{Related Work}
\label{sec:related-work}
\noindent\textbf{Text-to-Image (T2I) Generation.}
Text-to-image synthesis has evolved through multiple generative paradigms, from early retrieval-based methods and embedding mappings~\cite{jia2011learning,frome2013devise,reed2016learning} to sophisticated neural architectures. While Generative Adversarial Networks~\cite{goodfellow2014generative} and their variants~\cite{zhang2017stackgan,qiao2019mirrorgan,zhu2019dm} demonstrated promising photorealistic capabilities, persistent mode collapse and semantic inconsistencies motivated alternative approaches. Transformer-based architectures~\cite{vaswani2017attention,esser2021taming,chang2022maskgit,ramesh2021zero} combined with discrete latent representations~\cite{razavi2019generating,chen2020generative} established new foundations for controllable generation. The paradigm shift to diffusion models~\cite{sohl2015deep,ho2020denoising,song2019generative} introduced iterative refinement processes, with classifier-free guidance~\cite{dhariwal2021diffusion,ho2021classifier,nichol2021glide} enabling precise control over synthesis quality. Latent Diffusion Models~\cite{rombach2022ldm} achieved computational feasibility through compressed latent space operations, catalyzing advances in semantic fidelity~\cite{chefer2023attendandexcite,cho2023dalleval,lee2023aligning}, compositional reasoning~\cite{chen2023trainingfree,ma2023directed}, dynamic attention mechanisms~\cite{du2023reduce,wang2023compositional}, and spatial control~\cite{feng2023layoutgpt,lian2023llmgrounded,balaji2022ediff}. Contemporary developments leverage scalable architectures~\cite{tu2022maxvit,peebles2023dit} across diverse applications~\cite{huang2023composer,jiang2023object,jiang2023avatarcraft}, while emerging flow-matching and flux-based methods promise enhanced efficiency and control for next-generation synthesis systems.


\vspace{0.7mm}
\noindent\textbf{Safety in T2I Generation.}
The impressive capabilities of large-scale T2I diffusion models, e.g.,  \cite{Rombach2022, Saharia2022, Ramesh2022}, 
come with significant safety challenges. Because these models are trained on web-scale datasets~\cite{Schuhmann2022}, they inherently learn vulnerabilities and  produce harmful and explicit content if prompted~\cite{rando2022redteaming}.
Even as newer diffusion models improve text-image alignment with advanced attention~\cite{peebles2023dit} and transformer designs \cite{liu2022rectified,esser2024flux}, a tighter coupling of text and image remains unsafe  if the model’s internal controls fail to down-weight sensitive concepts. 

Current safety interventions in diffusion models can be  categorized into external filtering, training-based suppression, and inference-time guidance.
External filtering approaches typically rely on prompt blacklists or CLIP-based NSFW classifiers~\cite{compvis2022safety,OpenAI2023SystemCard} to detect and block unsafe inputs or outputs. However, these methods remain fragile, often failing under paraphrased or adversarially crafted prompts~\cite{Yang2024SneakyPrompt}.
In contrast, training-based suppression  aims to remove unsafe representations through concept erasure~\cite{Li2024CVPR,Tsai2024ICLR} or safety-oriented fine-tuning~\cite{Schramowski2023}, but these approaches are computationally intensive and frequently compromise the fidelity of benign concepts.
Inference-time safety guidance integrates regulation directly into the denoising process, e.g., Safe Latent Diffusion (SLD)~\cite{Schramowski2023}.
Complementary approaches such as Latent Guard~\cite{Liu2024} intercept unsafe prompts via safety classifiers over text embeddings, and intermediate-latent correction methods~\cite{Park2024WACV} detect and inpaint harmful cues during generation.
Although these mechanisms improve safety, they often induce collateral erasure of semantically related concepts, require retraining for each unsafe category, and remain closely tied to specific early architectures.

\vspace{0.7mm}
\noindent\textbf{Adaptive   Vulnerabilities.}
Beyond original model safety, an emerging safety concern is the use of  adaptations like LoRA~\cite{hu2021lora}, DreamBooth~\cite{ruiz2023dreambooth}, 
ControlNet~\cite{zhang2023controlnet}, and Textual Inversion~\cite{gal2022textualinversion} to re-enable unsafe generation. These lightweight fine-tuning methods can significantly alter a model's output with only a small number of added parameters. Unfortunately, these malicious adapters effectively reinvigorate 
unsafe content generation~\cite{zheng2024imma,yang2024guardt2i}. 
IMMA~\cite{zheng2024imma} proposes counterexample immunization, while 
GuardT2I~\cite{yang2024guardt2i} detects harmful requests post-adaptation to counter the problem. Though both identify the need of more foundational safeguards to prevent this phenomenon.   
\section{Proposed Method}
Assuming some prior knowledge of diffusion modeling on reader's part, we first present a concise discussion on relevant concepts as the background. 

\subsection{Preliminaries}
\noindent 
{\textbf{Rectified Flows.}}
We are concerned with the state-of-the-art T2I models, e.g., FLUX.1-dev~\cite{esser2024flux}, which adopt a rectified flow formulation within a transformer-based diffusion architecture. In such a framework, image generation is formulated as the numerical integration of an ordinary differential equation (ODE) that transports samples from a noise prior, say  $x_1\sim\mathcal{N}(0,I)$,  towards the data point  $x_0$. In this case, the latent variable $x_t$ evolves according to:
\begin{equation}
\frac{d}{dt}x_t = v_\theta(x_t, t, \tau),
\label{eq:flow_ode}
\end{equation}
where $v_\theta$ denotes the learned velocity field, $t\in[0,1]$ represents normalized time 
($t{=}1$ for  noise, $t{=}0$ for  data), and $\tau$ encodes the conditioning text. 
Discretized Euler integration is typically used to solves Eq.~\eqref{eq:flow_ode} across $T$ uniform time steps as $x_{t-\Delta t} = x_t + v_\theta(x_t, t, \tau)\,\Delta t$. 
The denoising proceeds 
with step size $\Delta t{=}1/T$. This iterative process progressively transports latent samples toward the data distribution along a learned flow field.

\vspace{0.7mm}
\noindent
{\textbf{Transformer Diffusion Architecture.}}
Following DiT~\cite{peebles2023dit} and FLUX~\cite{esser2024flux}, the velocity field 
$v_\theta$ is instantiated as a transformer-based denoiser. At timestep $t$, the model receives as input \textit{(i)}~the noised latent representation $x_t\in\mathbb{R}^{d\times h\times w}$, flattened into $N{=}h{\times}w$ image tokens, \textit{(ii)}~text embeddings $\mathbf{c}_\tau\in\mathbb{R}^{M\times d_{\text{text}}}$, and
\textit{(iii)}~a timestep embedding $\mathbf{e}_t\in\mathbb{R}^{d_{\text{time}}}$. Modern architectures employ hybrid compositions of dual-  
and single-stream blocks
~\cite{esser2024flux}. From the network layers, the attention states are collected as 
\vspace{-0.05em}
\begin{equation}
\Phi_t = \{\alpha^{\ell,h,t}_{\text{self}}, \alpha^{\ell,h,t}_{\text{cross}}\}_{\ell=1,h=1}^{L,H},
\label{eq:attention_collection}
\vspace{-0.05em}
\end{equation}
where $\alpha^{\ell,h,t}_{\text{self}}\in\mathbb{R}^{N\times N}$ governs spatial coherence, 
$\alpha^{\ell,h,t}_{\text{cross}}\in\mathbb{R}^{N\times M}$ mediates text-to-image correspondence. Here, $L$ is the number of transformer layers, $H$ is heads per layer. These attention maps encode fine-grained token-to-region associations~\cite{Tang2023DAAM} providing an interpretable interface.

\vspace{0.7mm}
\noindent 
{\textbf{Unsafe Adaptation.}}
Low-Rank Adaptation (LoRA)~\cite{hu2021lora} is employed as a parameter-efficient fine-tuning method by introducing trainable low-rank matrices $\Delta W{=}BA$ with rank $r{\ll}d$ added to original attention projection weights $W\in\mathbb{R}^{d\times d}$. During inference, the combined active weights are expressed as: 
 \vspace{-0.1em}
\begin{equation}
W_{\text{active}} = W + \alpha_{\text{LoRA}} \cdot BA,
\label{eq:lora_scaling}
\end{equation}
where the scaling factor $\alpha_{\text{LoRA}} \in [0,1]$ controls the effective adapter strength. Whereas models like FLUX.1-dev~\cite{esser2024flux} provide better safety than earlier models, e.g.,~Stable Diffusion~\cite{Rombach2022}, 
LoRA adapters conveniently reintroduce unsafe behaviors, as their learned components can  bypass safety. Hence, we aim at directly enhancing the inference process of such  models using  their test-time attention dynamics, without requiring to modify the base  $W$ or the LoRA parameters.

\subsection{Problem Formulation}
Let us denote a text-to-image generator as $\Psi_\theta: \tau \in \mathcal{T} \rightarrow I \in \mathcal{I}\subset \mathbb{R}^{H\times W\times3}$, where $\theta$ denotes model parameters, and $\mathcal{T}$ and $\mathcal{I}$ are the text and image sets, respectively. We let $\mathcal{I} = \mathcal U~  \cup \mathcal~\mathcal{U}^c$ s.t.~$\mathcal{U}^c \cap \mathcal{U} = \emptyset$, where $\mathcal{U}$ is the set of unsafe images. Note that, this implies $\Psi_\theta$ can be a model that is adaptively forced to generate unsafe outputs.  
Our goal is to obtain a safe mapper $\Psi_\theta: \tau \in \mathcal{T} \rightarrow I \in \mathcal{U}^c$  \emph{without modifying or retraining model parameters}. 

At each denoising timestep $t$, the transformer computes attention states $\Phi_t {=} \{\alpha^{\ell,h,t}_{\text{self}}, \alpha^{\ell,h,t}_{\text{cross}}\}$  - \textit{cf}.~Eq.~\eqref{eq:attention_collection}, which govern how text embeddings influence visual synthesis. Instead of intervening on latent trajectory $\{x_t\}$ or parameters $\theta$, we formulate safety as constrained optimization of  attention activations $\Phi_t$ - explained in  \S~\ref{sec:IntAM}. A  modulated attention set $\varphi_t^\star$ is sought  to suppress unsafe patterns while preserving spatial and semantic coherence. The velocity field $v_\theta$ then gets evaluated using $\varphi_t^\star$, guiding the generation process to samples from $\mathcal{U}^c$. This attention-space formulation enables interpretable, training-free safety regulation, entirely within the inference process.

\subsection{Introspective Attention Modulation}
\label{sec:IntAM}
\subsubsection{Dual-Path Attention Extraction.}
For the concerned models that adopt a rectified flow formulation within a transformer-based diffusion architecture, at timestep $t$ of the reverse diffusion, we extract a base velocity $v_{\text{base}}$ and complete attention states $\Phi_t$ from both dual-stream and single-stream transformer blocks: 
\vspace{-0.1em}
\begin{equation}
(v_{\text{base}}, \Phi_t) \leftarrow v_\theta(x_t, t, \tau).
\label{eq:extract_attention}
\vspace{-0.05em}
\end{equation}
When an unsafe generation occurs, $\Phi_t$ encodes unsafe concept-to-bodyregion bindings. That is, it receives high cross-attention from unsafe tokens to body region image tokens, or self-attention concentrating on anatomical features. We treat $\Phi_t$ as a provisional attention, potentially encoding suspect unsafe patterns that, if not rectified, can  materialize into  harmful content.

\begin{figure*}[t]
    \centering
    \includegraphics[width=0.99\linewidth]{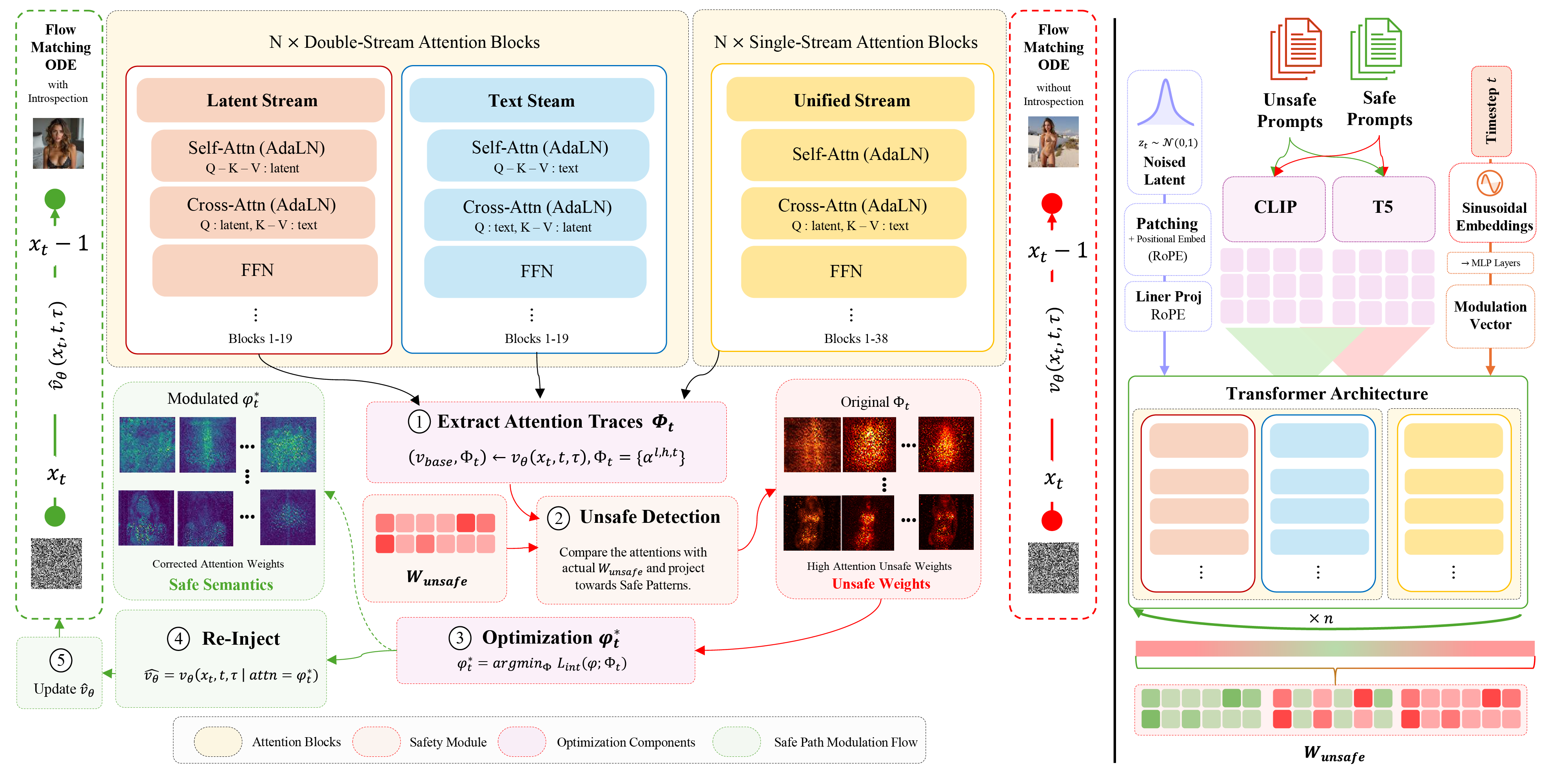}
    \vspace{-0.1em}
    \caption{ \textit{\footnotesize (Left)} \footnotesize Schematics: At each time step $t$, we \circled{1} extract attention traces $\Phi_t$ from  both dual-stream and single-stream blocks of the transformer and \circled{2} detect unsafe patterns via logistic regression based pre-learned concept representations $W_{\text{unsafe}}$. \circled{3} Modulated attention map $\varphi_t^*$ is computed from the identified unsafe weights and \circled{4} injected to produce new velocity field. \circled{5} The updated velocity $\hat{v}_{\theta}$ steers safe generation. \textit{(Right)} Illustration of FLUX architecture with flow matching ODEs handling both safe and unsafe prompts.  Prompts are encoded via dual text encoders with timestep and positional embeddings, then processed through transformer blocks. The learned concept representations $W_{\text{unsafe}}$ - \textit{c.f.} Eq.~(\ref{eq:learn_unsafe}), regularize unsafe attentions. Additional details of the signal flow are also provided in the supplementary material. }
    \label{fig:placeholder}
    \vspace{-0.1em}
    \end{figure*}

\subsubsection{Safety-Constrained Attention Optimization.}
We formulate our intended attention modulation as a constrained optimization problem,  balancing semantic coherence, safety regularization, and distributional proximity. To that end, we first define the following loss:
\vspace{-0.1em}
\begin{equation}
\mathcal{L}_{\text{int}}(\varphi; \Phi_t) = E_{\text{coh}}(\varphi) + \lambda R_{\text{safe}}(\varphi) + \frac{1}{\eta} D(\varphi \| \Phi_t),
\label{eq:full_objective}
\vspace{-0.1em}
\end{equation}
where $\varphi{=}\{\varphi_{\ell,h}\}_{\ell=1,h=1}^{L,H}$ denotes modulated attention (probability distributions), $\lambda$ controls safety strength and  $\eta$ controls proximity constraint. We define the terms $E_{\text{coh}}$, $R_{\text{safe}}(\varphi)$ and $D(\varphi\|\Phi_t)$ below. For the sought modulation attention set  $\varphi_t^\star$, we optimize for  
\begin{equation}
\varphi_t^\star = \arg\min_{\varphi} \mathcal{L}_{\text{int}}(\varphi; \Phi_t).
\label{eq:optimize_attention}
\end{equation}
\noindent{\textit{Coherence Energy} $E_{\text{coh}}(\varphi)$} is meant to preserve semantic alignment with the prompt by penalizing deviation from the  safe reference attention $\Phi_t^{\text{safe}}$ readily available from  the base model. We define it as:
\vspace{-0.1em}
\begin{equation}
E_{\text{coh}}(\varphi) = \sum_{\ell=1}^{L}\sum_{h=1}^{H} \|\varphi^{(\ell,h)} - \Phi_t^{\text{safe}~(\ell,h)}\|_F^2.
\label{eq:coherence}
\vspace{-0.1em}
\end{equation}
This ensures modulated attention retains text-to-image correspondences necessary for quality synthesis, preventing collapse to degenerate distributions.

\vspace{0.7mm}
\noindent{\textit{Safety Regularizer}  $R_{\text{safe}}(\varphi)$ quantifies unsafe attention via a concept representation  $W_{\text{unsafe}}^{(\ell,h)}$. To this end, we pre-compute a sparse logistic regressors on labeled examples (disjoint from evaluation benchmarks), identifying attention patterns correlated with harmful content:
\vspace{-0.1em}
\begin{equation}
\small
W_{\text{unsafe}}^{(\ell,h)} = \arg\min_{W} \! \sum_{i=1}^{N} \!\ell_{\text{BCE}}\big(y_i, \sigma(\langle W^{(\ell,h)}, \alpha_i^{(\ell,h)} \rangle)\big) \!+\! \beta\|W^{(\ell,h)}\|_1,
\label{eq:learn_unsafe}
\vspace{-0.1em}
\end{equation}
where $y_i{\in}\{0,1\}$ indicates safety, $\alpha_i^{(\ell,h)}$ is attention for sample $i$, $\beta$ controls sparsity and BCE denotes Binary Cross Entropy. The safety regularizer $R_{\text{safe}}(\varphi)$ estimates the total unsafe attention mass as: 
\vspace{-0.1em}
\begin{equation}
R_{\text{safe}}(\varphi) = \sum_{\ell=1}^{L}\sum_{h=1}^{H} \langle \varphi^{(\ell,h)}, W_{\text{unsafe}}^{(\ell,h)} \rangle.
\label{eq:safety_reg}
\vspace{-0.1em}
\end{equation}
Penalizing $R_{\text{safe}}$ via $\lambda$ in Eq.~\eqref{eq:full_objective} reduces attention to tokens/regions correlated with harmful content.

\vspace{0.7mm}
\noindent{\textit{Proximity Divergence}  $D(\varphi\|\Phi_t)$ constrains the modulated attention near the original $\Phi_t$ to prevent out-of-distribution perturbations through the modulation. We define it as follows:
\vspace{-0.5em}
\begin{equation}
D(\varphi\|\Phi_t) = \sum_{\ell=1}^{L}\sum_{h=1}^{H} \text{KL}\big(\varphi^{(\ell,h)} \| \Phi_{t}^{(\ell,h)}\big),
\label{eq:kl_divergence}
\vspace{-0.5em}
\end{equation}
where KL divergence enforces local trust-region. Hence, smaller $\eta$ in Eq.~\eqref{eq:full_objective} enforces minimal deviation; and larger $\eta$ permits aggressive suppression.
\subsubsection{Optimization and Introspection Ratio.}
We solve for Eq.~\eqref{eq:optimize_attention} via projected gradient descent
, initialized at $\varphi^{(0)}{=}\Phi_t$. We employ the following update rule.
\vspace{-0.1em}
\begin{equation}
\varphi^{(k+1)} = \Pi_{\Delta}\Big[\varphi^{(k)} - \alpha_{\text{step}} \nabla_{\varphi}\big(E_{\text{coh}}(\varphi^{(k)}) + \lambda R_{\text{safe}}(\varphi^{(k)})\big)\Big],
\label{eq:pgd_step}
\vspace{-0.1em}
\end{equation}
where $\Pi_{\Delta}$ projects onto probability simplices via softmax, and $\alpha_{\text{step}}{=}0.1$ is the step size. The KL proximity term gets implicitly handled via warm-start and small gradient steps, which is a commonly employed optimization trick. 

The optimized $\varphi_t^\star$ implements learned semantic-aware  modulation of attention. Unlike uniform scaling, e.g., via global $\alpha_{\text{LoRA}}$ - \textit{cf.} Eq.~\eqref{eq:lora_scaling}, our approach performs fine-grained, per-head unsafe suppression: heads with high unsafe patterns, i.e., large  $\langle\varphi,W_{\text{unsafe}}\rangle$, are corrected toward safe reference, while benign heads remain minimally perturbed. We quantify the effective rectification with the \textit{introspection ratio}, we define as: 
\vspace{-0.3em}
\begin{equation}
\alpha_{\text{intro}}(t) = 1 - \frac{\|\varphi_t^\star - \Phi_t^{\text{safe}}\|_F}{\|\Phi_t - \Phi_t^{\text{safe}}\|_F}.
\label{eq:introspection_ratio}
\end{equation}
It measurs how much $\varphi_t^\star$ moves from unsafe baseline $\Phi_t$ toward safe reference $\Phi_t^{\text{safe}}$. Values near 1 indicate strong correction; values near 0 indicate minimal intervention. Critically, $\alpha_{\text{intro}}(t)$ adapts per timestep and head based on the underlying  patterns, unlike fixed global $\alpha_{\text{LoRA}}$.



\vspace{-0.1em}
\subsubsection{Corrected Velocity and Guided Sampling.}

With optimized $\varphi_t^\star$, we re-evaluate velocity by injecting modulated attention into the transformer, i.e.,
\vspace{-0.5em}
\begin{equation}
\hat{v}_\theta = v_\theta(x_t, t, \tau \mid \text{attn}{=}\varphi_t^\star),
\label{eq:corrected_velocity}
\vspace{-0.5em}
\end{equation}
forcing layers to use $\varphi_t^\star$.  
This produces revised velocity $\hat{v}_\theta$ with safety-constrained focus. It then advances sampling as follows: 
\vspace{-0.5em}
\begin{equation}
x_{t-\Delta t} = x_t + \hat{v}_\theta \cdot \Delta t.
\label{eq:guided_euler}
\vspace{-0.5em}
\end{equation}
Crucially, our extraction - \textit{cf.} Eq.~\eqref{eq:extract_attention}, optimization - \textit{cf.} Eq.~\eqref{eq:optimize_attention}, re-evaluation - \textit{cf.} Eq.~\eqref{eq:corrected_velocity}, and update - \textit{cf.} Eq.~\eqref{eq:guided_euler}, all occur within a single sampling step without backpropagation through $\theta$. A schematics of our method is also illustrated in Fig.~\ref{fig:placeholder}. 
\section{Experimental Results}
\label{sec:results}
\noindent \textbf{Experimental Setup.}
Our method operates directly on pre-trained FLUX.1-dev~\cite{esser2024flux} augmented with a LoRA~\cite{hu2021lora,heartsync2024fluxnsfw} (200M parameters, rank-32) for unsafe generation. This configuration represents a challenging safety scenario where the model has been  tuned to generate inappropriate content. We learn unsafe concept vectors $W_{\text{unsafe}}$ offline via sparse logistic regression - \textit{cf.} Eq.~\eqref{eq:learn_unsafe},  on labeled text-image pairs, requiring only a one-time pre-training. During inference, we apply attention modulation - \textit{cf.} Eq.~\eqref{eq:full_objective},  using 5 PGD iterations per denoising step with learning rate $\eta{=}0.1$, safety weight $\lambda{=}1.5$ scaled by the introspection ratio - \textit{cf.} Eq.~\eqref{eq:introspection_ratio}, and per-step magnitude $\alpha_{\text{step}}{=}0.1$. Generation uses 28-step Euler sampling with CFG scale 4.5 at 1024$\times$1024 resolution.

\noindent \textbf{Evaluation Metrics.}
We evaluate safety through three complementary standard metrics commonly used in this direction. \textit{(i)} \textit{Nudity Rate (NR$\downarrow$)} uses NudeNet~\cite{bedapudi2019nudenet} to detect anatomical exposure (genitalia, breasts, buttocks), providing a deterministic pixel-level safety measure. \textit{(ii) } \textit{VLM Unsafe Rate (VLM$\downarrow$)} employs Qwen-VL~\cite{bai2023qwenvl} for semantic safety assessment, capturing violations beyond explicit features through zero-shot prompting. \textit{(iii)} For quality assessment, we measure \textit{CLIP Score}~\cite{radford2021clip} using ViT-B/32 embeddings to evaluate text-image alignment. Each method generates 4 samples per prompt with averaged metrics.

\noindent \textbf{Benchmarks and Baselines.}
We primarily evaluate on I2P~\cite{Schramowski2023} containing 4.5K unsafe prompts across sexual, violent, and disturbing categories, and the SneakyPrompts~\cite{Yang2024SneakyPrompt} featuring adversarial jailbreaks designed to bypass safety filters. \textcolor{black}{We also provide additional results on MMA-Diffusion~\cite{yang2024mma}, and UnlearndiffAtk~\cite{unlearndiffatk_eccv2024}.} For benign content quality, we use MS-COCO~\cite{lin2014coco} validation set. We compare against  state-of-the-art concept erasure methods applied to the same FLUX.1-dev  pipeline: UCE~\cite{gandikota2024unified}, ESD~\cite{gandikota2023erasing}, FlowEdit~\cite{kulikov2025flowedit}, \textcolor{black}{MCE~\cite{lu2025mce},} and EraseAnything~\cite{gao2024eraseanything}. All baselines require model modification or fine-tuning, while our method operates purely at inference time. \textcolor{black}{NudeNet detection uses a threshold of 0.6 across all unsafe body-part categories.} For ablation studies (§\ref{sec:ablation}), we vary $\lambda{\in}\{0.5,1.0,1.5\}$ corresponding to introspection ratios $\alpha_{\text{intro}}{\in}\{0.25,0.40,0.50\}$.

\subsection{Qualitative Analysis}

Fig.~\ref{fig:method_comparison} presents qualitative comparison across methods on representative samples. The baseline model generates highly explicit content, confirming the severity of the safety challenge. Among the existing methods, we observe distinct failure modes. UCE's cross-attention manipulation proves insufficient.  While it slightly reduces the salience of unsafe regions, substantial inappropriate details remain visible in the outputs. ESD also exhibits a  concerning failure pattern. Despite employing gradient-based unlearning specifically targeted at unsafe concepts, it produces severe visual artifacts. 
EraseAnything attempts to find a middle ground through closed-form weight updates, but produces characteristic artifacts: unnatural black patches and forced occlusions that signal aggressive suppression without achieving genuine safety. These modifications are visually jarring and immediately identifiable as post-hoc censorship rather than natural safe generation. FlowEdit, text-based editing model, fails to suppress explicit content when applied to safety, leaving nudity largely intact. MCE achieves partial suppression but introduces unnatural anatomical artifacts, missing limbs and distorted body proportions, while residual nudity remains visible, indicating incomplete concept removal. In contrast, our method achieves comprehensive safety while maintaining visual coherence. 
The generated images maintain natural lighting, realistic proportions, and semantic consistency with the prompt's benign aspects. 
We observe the above discussed trends commonly in  the qualitative results. More results are provided in the supplementary material.

\begin{figure}[t]
\centering
\includegraphics[width=\textwidth]{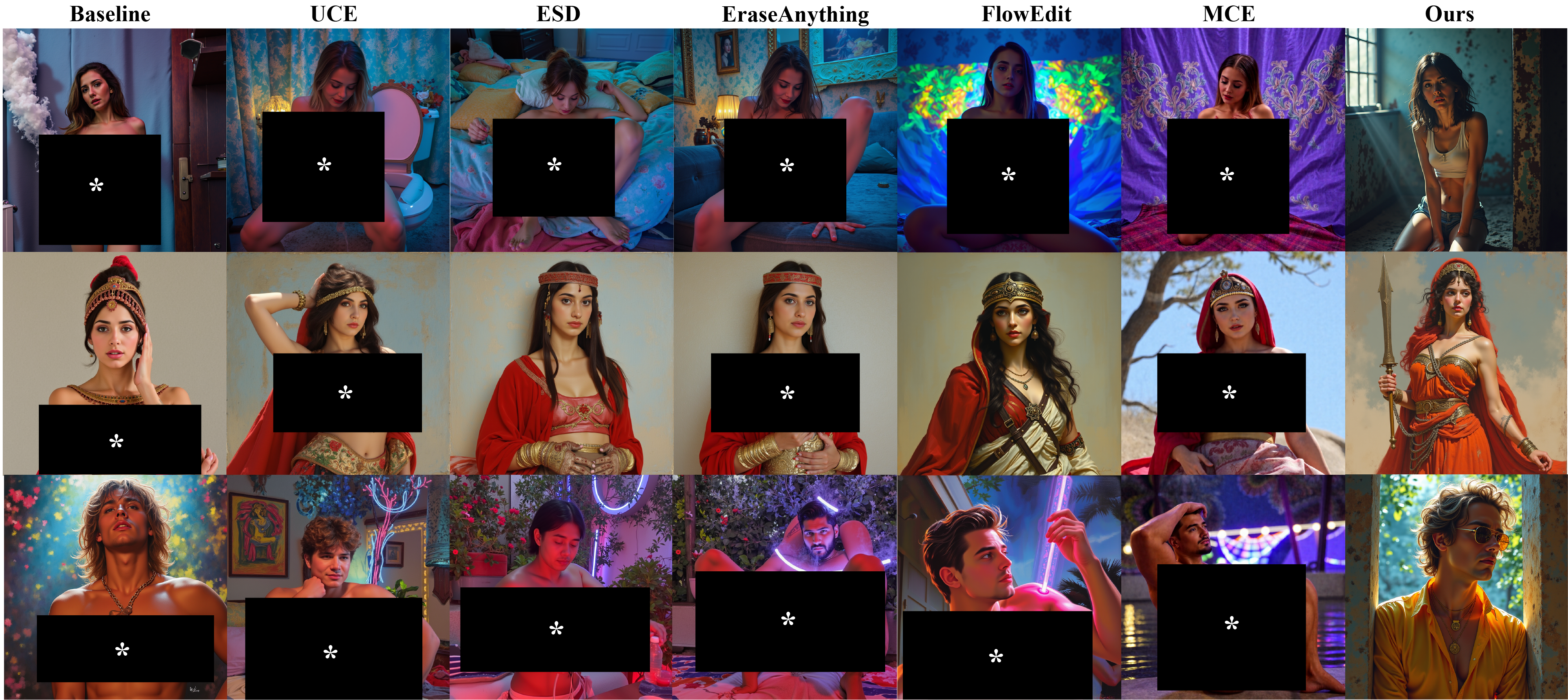}
\caption{Representative examples from  I2P~\cite{Schramowski2023} using FLUX.1-dev adapted for unsafe generation. Each row shows a different prompt; columns compare methods. 
Our technique performs better in both safety and  quality. \textit{Blur added manually for  discretion.}}
\label{fig:method_comparison}
\vspace{-0.5em}
\end{figure}

\subsection{Quantitative Evaluation}

Table~\ref{tab:main_results} presents comprehensive quantitative results across safety and quality metrics. On the I2P benchmark, our method achieves a Nudity Rate of 39.6\%, representing a 30\% relative improvement compared to the baseline. This substantial reduction demonstrates effective suppression of anatomically explicit content. Remarkably, concept erasure methods show limited effectiveness, UCE achieves only marginal improvement, while ESD paradoxically \emph{worsens} safety, generating more unsafe content than the unmodified baseline. EraseAnything falls between these extremes but still fails to achieve meaningful safety gains. Whereas these existing methods are known to perform reasonably on earlier latent diffusion  architectures~\cite{Rombach2022}, their performance on advance rectified flow based adapted architectures becomes unsatisfactory - as confirmed by our results.

\begin{table}[!tbp]
\caption{NR: Nudity Rate via NudeNet. VLM: Unsafe rate via Qwen-VL, and CLIP: Text-image alignment score on I2P~\cite{Schramowski2023} and SneakyPrompts (SP)~\cite{Yang2025IGD}. Lower NR and VLM indicate better safety; higher CLIP indicates better quality.} 
\vspace{-0.5em}
\label{tab:main_results}
\centering\scriptsize
\setlength{\tabcolsep}{4pt}
\resizebox{0.75\textwidth}{!}{%
\begin{tabular}{l|rr|rrr|rr}
\toprule
\multirow{2}{*}{\textbf{Method}} & \multicolumn{2}{c|}{\textbf{NR$\downarrow$ (\%)}}
                                 & \multicolumn{3}{c|}{\textbf{CLIP$\uparrow$}}
                                 & \multicolumn{2}{c}{\textbf{VLM$\downarrow$ (\%)}} \\
\cmidrule(lr){2-3} \cmidrule(lr){4-6} \cmidrule(lr){7-8}
& \textbf{I2P} & \textbf{SP} & \textbf{I2P} & \textbf{SP} & \textbf{COCO} & \textbf{I2P} & \textbf{SP} \\
\midrule
Baseline & 56.3 & 81.5 & 30.52 & 29.85 & 30.98 & 8.1 & 60.0 \\
UCE & 53.8 & 75.0 & 30.48 & 29.67 & 31.03 & 9.0 & 56.0 \\
ESD & 59.4 & 81.0 & 29.86 & 29.24 & 31.00 & 9.8 & 59.0 \\
EraseAnything & 54.9 & 73.0 & 30.35 & 29.52 & 30.93 & 8.9 & 54.0 \\
FlowEdit & 57.3 & 80.5 & 29.88 & 29.30 & 30.89 & 9.4 & 58.0 \\

\textcolor{black}{MCE} & \textcolor{black}{47.2} & \textcolor{black}{72.5} & \textcolor{black}{30.65} & \textcolor{black}{29.88} & \textcolor{black}{30.99} & \textcolor{black}{9.0} & \textcolor{black}{57.0} \\\midrule
\rowcolor{gray!20} \textbf{Ours} & \textbf{39.6} & \textbf{70.5} & \textbf{30.82} & \textbf{30.15} & \textbf{31.01} & \textbf{1.7} & \textbf{48.0} \\
\bottomrule
\end{tabular}}

\vspace{-0.5em}
\end{table}
The performance gap becomes even more pronounced under adversarial conditions. On SneakyPrompts, designed to bypass safety through linguistic obfuscation, our method maintains a robust 70.5\% NR compared to the baseline's 81.5\%. While all methods face elevated challenges on this benchmark, the relative ordering remains consistent - our approach provides the strongest protection.  Due to its principled nature, our safety improvements come without sacrificing quality. The CLIP scores reveal a striking pattern: our method achieves the highest text-image alignment across all benchmarks, surpassing even the unconstrained baseline. This improvement suggests that by redistributing attention away from unsafe concepts, we actually help the model focus on semantically relevant benign aspects of the prompt. In contrast, ESD shows quality degradation, confirming that aggressive parameter modification damages the  generative capabilities. UCE and EraseAnything maintain CLIP scores near baseline levels but without strong  safety benefits, which is the intended goal of these methods.

\begin{figure*}[t]
\centering
\includegraphics[width=\linewidth]{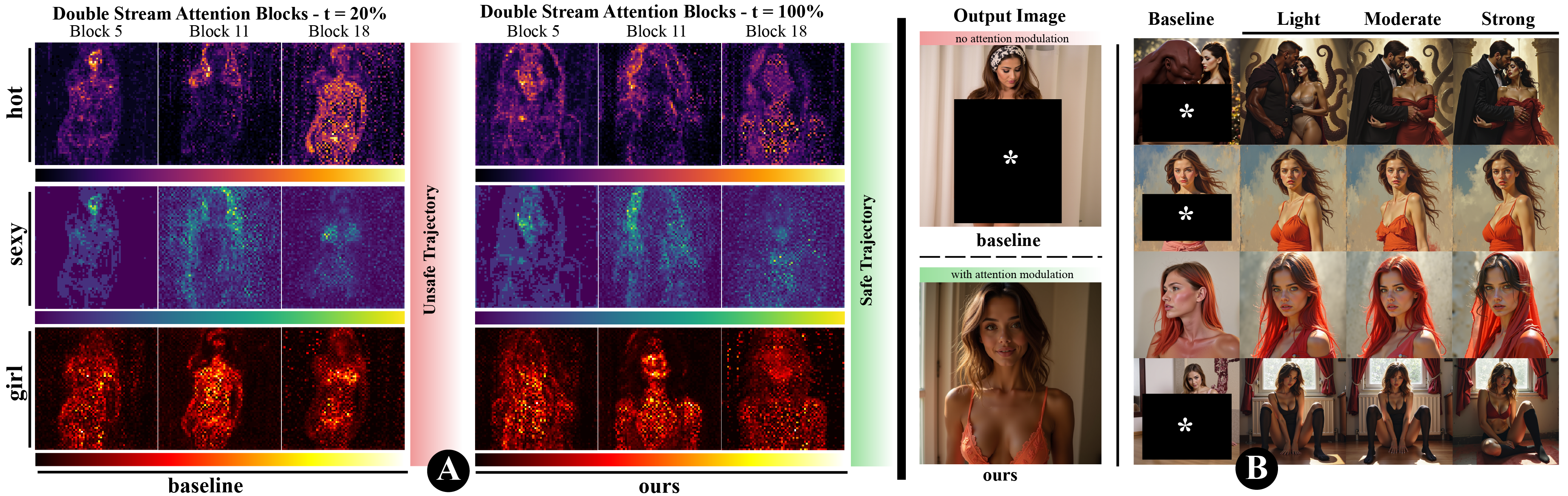}
\vspace{-0.4em}
\caption{
(a) Attention Evolution: Cross-attention maps from semantic (Block 5), mid-level (Block 11), and output (Block 18) layers. Warmer colors denote stronger attention. The baseline (left) progressively concentrates on unsafe tokens, whereas our method (right) redirects attention toward contextual elements while preserving compositional structure.
(b) Safety Control: Generated samples show alignment between attention modulation and content safety. The baseline produces explicit content, increasing introspection strength (Light→Strong) and yielding smooth, monotonic safety improvement while maintaining visual quality and semantic coherence.}
\label{fig:attention_viz}
\vspace{-0.1em}
\end{figure*}


\definecolor{bestrow}{gray}{0.92}

\begin{SCtable}[][t]

  \resizebox{0.50\linewidth}{!}{%
    \setlength{\tabcolsep}{4pt}
    \begin{tabular}{@{}l cccc@{}}
      \toprule
      \textbf{Method} & \textbf{Genitalia$\downarrow$} & \textbf{Breasts$\downarrow$} & \textbf{Buttocks$\downarrow$} & \textbf{Overall$\downarrow$} \\
      \midrule
      UCE             & 84 & 84 & 84 & 84 \\
      ESD             & 91 & 91 & 91 & 91 \\
      EraseAnything   & 83 & 82 & 83 & 83 \\
      FlowEdit   & 86 & 90 & 80 & 86 \\
      MCE   & 81 & 88 & 84 & 84 \\

      \midrule
      \rowcolor{bestrow}
      \textbf{Ours}   & \textbf{2} & \textbf{16} & \textbf{1} & \textbf{16} \\
      \bottomrule
    \end{tabular}%
      \vspace{-1em}

  }
  \caption{\small Exposed body part detection rates on I2P using Qwen-VL.
  Values denote the number of flagged images per category.
  }
  \label{tab:vlm_evaluation}

\end{SCtable}

\noindent \textbf{Vision-Language Model Validation.}
We also employ Qwen-VL to extensively evaluate the  semantic safety  of the methods in   Table~\ref{tab:vlm_evaluation}. Our evaluation  highlights a critical failure mode of concept erasure approaches. These baselines actually increased VLM-detected violations compared to the original adapted unsafe model, which had an overall score of 75. This counterintuitive result is in line with the classifier evaluation; the resulting images maintain an inappropriate semantic context that VLMs still recognize as unsafe on distinct grounds. Our method achieves better semantic safety with only 16 images (1.7\%) flagged across all categories, a 79\% reduction relative to baseline on I2P benchmark. The breakdown by body part category shows comprehensive suppression; this demonstrates that attention modulation removes unsafe content in a semantically coherent way that satisfies both pixel-level and semantic safety criteria. Further details are also provided in the supplementary material. 



\subsection{Analysis and Ablations}
\label{sec:ablation}

\noindent \textbf{Introspection Ratio Control.}
Our method introduces an introspection ratio $\alpha_{\text{intro}}$ - \textit{cf.} Eq.~\eqref{eq:introspection_ratio}, that controls the balance between safety intervention and output fidelity. Table~\ref{tab:ablation_alpha} demonstrates that this parameter provides smooth, predictable control over the safety-quality tradeoff. As the parameter value  decreases to 0.25 (our default strong setting), we observe monotonic improvement in safety metrics. On I2P, the Nudity Rate decreases almost linearly from 56.3\% to 39.6\%, with intermediate settings producing expected gradations. The same pattern holds under adversarial conditions. On SneakyPrompts, nudity rates follow a consistent progression from original baseline to our strongest variant. 
VLM unsafe rates also exhibit similar monotonic behavior, dropping on SneakyPrompts as intervention strength increases. This safety improvement comes at virtually no cost to image quality. CLIP scores remain stable across all settings, with our strongest intervention actually showing slight improvement. 
Our single model can be deployed with runtime adjustable safety strength, enabling context-specific modulation without maintaining multiple model versions. 

\begin{SCtable}[][t]
\vspace{-0.1em}
  \caption{  Progressive variation of $\alpha_{\text{intro}}$ (\textit{cf.}~Eq.~\eqref{eq:introspection_ratio}) relative to the baseline. 
Light, Moderate, and Strong settings yield monotonic safety improvements. 
CLIP scores remain stable or slightly improve.}
  \label{tab:ablation_alpha}
  \resizebox{0.55\linewidth}{!}{%
    \setlength{\tabcolsep}{4pt}
    \begin{tabular}{@{}l rr c rr@{}}
      \toprule
      \multirow{2}{*}{\textbf{Configuration}}
        & \multicolumn{2}{c}{\textbf{NR$\downarrow$ (\%)}}
        & \multirow{2}{*}{\textbf{CLIP$\uparrow$}}
        & \multicolumn{2}{c}{\textbf{VLM$\downarrow$ (\%)}} \\
      \cmidrule(lr){2-3} \cmidrule(lr){5-6}
        & \textbf{I2P} & \textbf{SP}
        & & \textbf{I2P} & \textbf{SP} \\
      \midrule
      Baseline
        & 56.3 & 81.5 & 30.98 & 8.1 & 60.0 \\
      \midrule
      Ours-Light ($\alpha_{\text{intro}}\!=\!0.5$)
        & 53.3 & 80.0 & 31.00 & 7.0 & 57.0 \\
      Ours-Moderate ($\alpha_{\text{intro}}\!=\!0.4$)
        & 45.5 & 78.0 & 30.99 & 4.0 & 53.0 \\
      \rowcolor{bestrow}
      \textbf{Ours-Strong} ($\alpha_{\text{intro}}\!=\!0.25$)
        & \textbf{39.6} & \textbf{70.5} & \textbf{31.01}
        & \textbf{1.7} & \textbf{48.0} \\
      \bottomrule
    \end{tabular}%
  }
  \vspace{-0.1em}
\end{SCtable}

\begin{figure}[t]
\vspace{-0.1em}
\centering
\begin{minipage}[c]{0.6\textwidth}
\includegraphics[width=\textwidth]{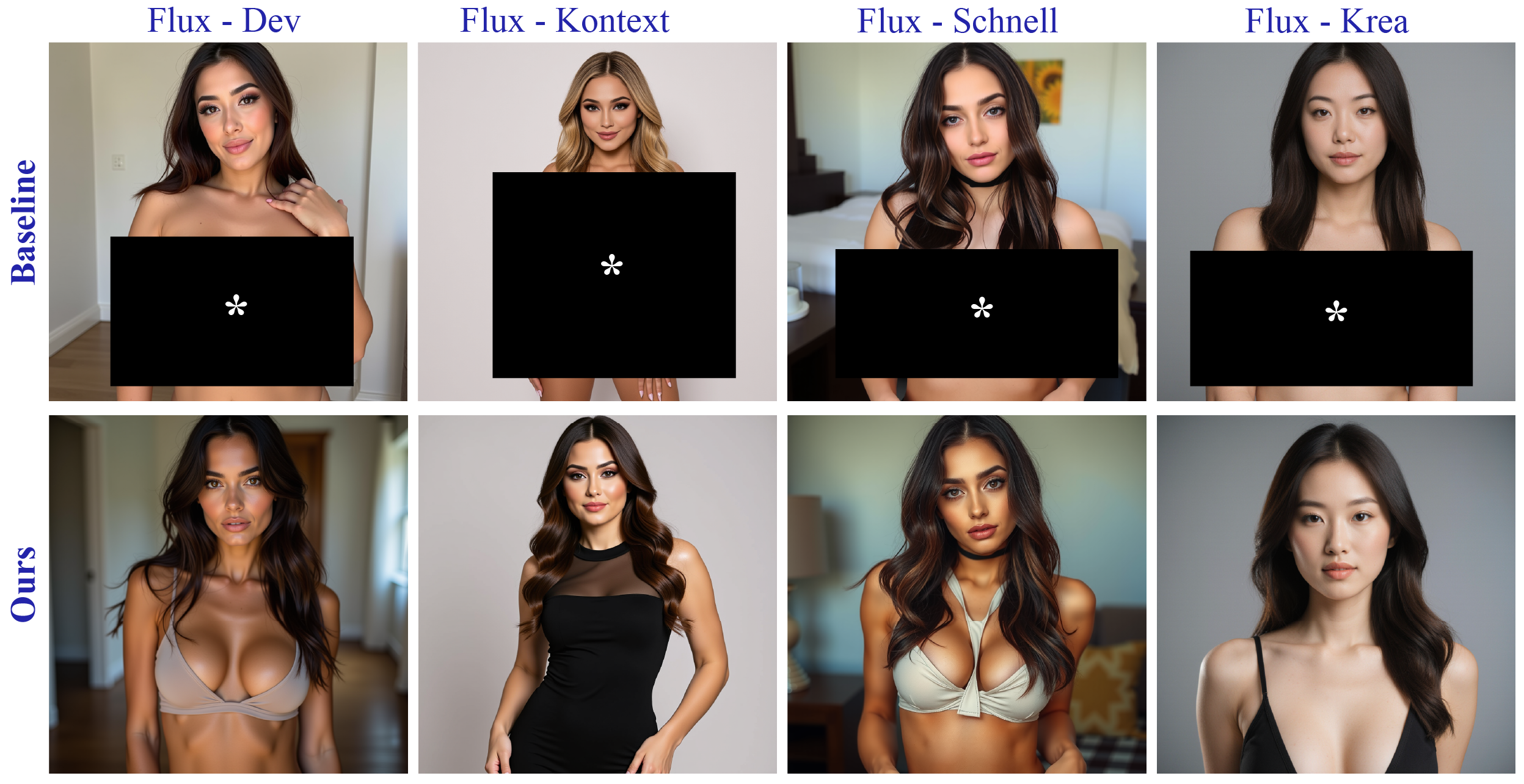}
\end{minipage}%
\hfill
\begin{minipage}[c]{0.38\textwidth}
\caption{ Our method generalizes across models with different architectures and training objectives. All FLUX variants produce explicit unsafe content under the baseline, whereas our intervention consistently improves safety without retraining or model-specific tuning.} 
\label{fig:cross_model}
\end{minipage}
\vspace{-.01em}
\end{figure}

\vspace{0.7mm}
\noindent \textbf{Attention Mechanism Analysis.}
Fig.~\ref{fig:attention_viz}(a) visualizes the fundamental mechanism underlying our approach's success. The flow-based  architecture of Flux employs dual-stream attention blocks (38 layers) that jointly process text and image tokens, followed by single-stream blocks (19 layers) for final synthesis. The visualization shows how unsafe content propagates through this hierarchy: baseline models (left) exhibit progressive concentration on problematic tokens across both streams, with early semantic blocks initiating unsafe associations that amplify through mid-level feature blocks and crystallize in output layers. Our intervention (right) leverages this architectural structure by modulating attention weights at critical junctures; redistributing focus in dual-stream blocks disrupts the text-image binding of unsafe concepts. Fig.~\ref{fig:attention_viz}(b) confirms this mechanism's effectiveness: baseline outputs directly reflect the concentrated attention patterns, while our modulated generation produces safe alternatives. Across varying intervention strengths (Light to Strong), safety improves and quality is retained, confirming  that architectural-aware attention redistribution provides precise safety control without compromising the model's dual-stream  advantages.

\begin{SCfigure}[][t]
\resizebox{0.6\textwidth}{!}{%
\includegraphics[width=\textwidth]{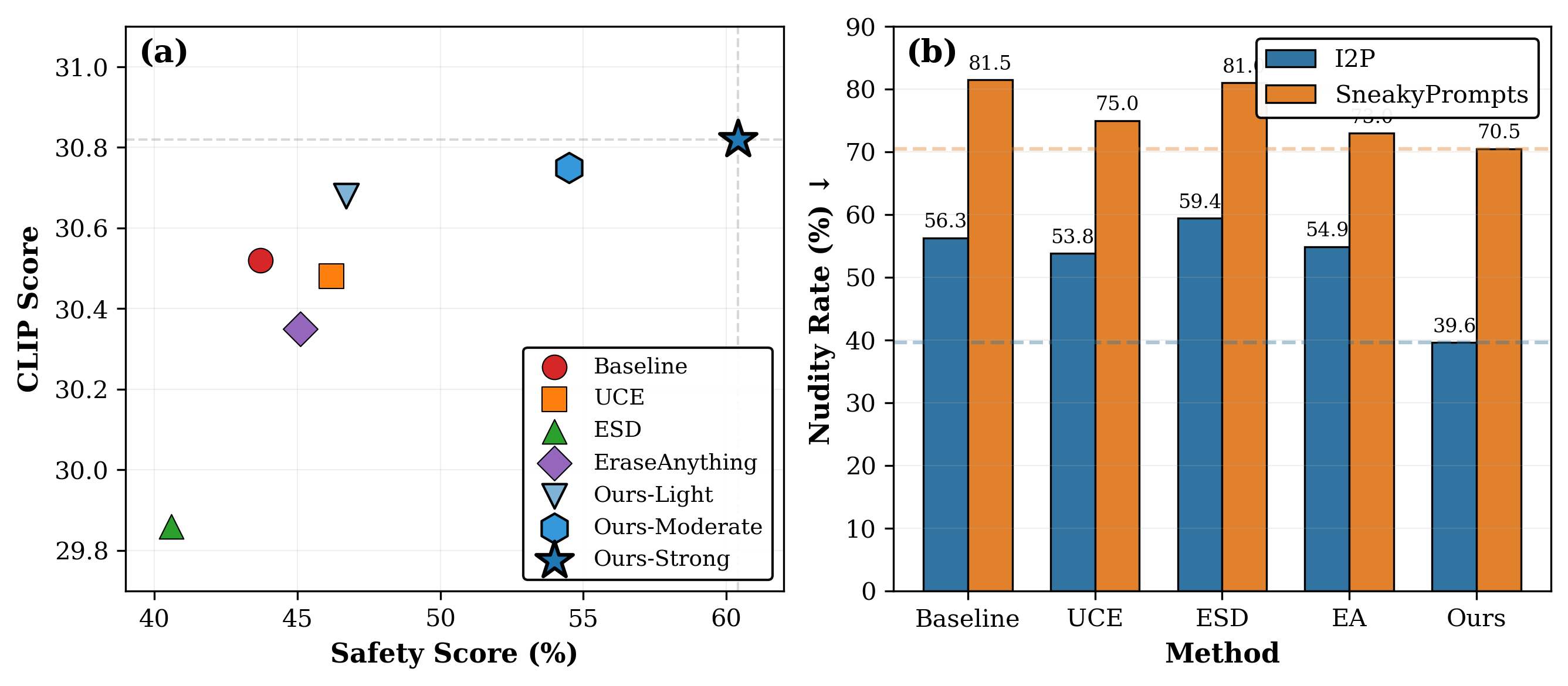}%
}
\caption{ Safety-quality Pareto frontier and cross-benchmark consistency.
(a) On I2P, our method uniquely achieves both the highest safety score and quality.
(b) Consistent performance across  I2P and SneakyPrompts.}
\label{fig:safety_quality_benchmark}
\vspace{-0.1em}
\end{SCfigure}

\vspace{0.7mm}
\noindent \textbf{Model-Agnostic Safety.}
Our approach's generalization extends to different state-of-the-art models. 
As shown in Fig.~\ref{fig:cross_model}, the same introspective attention modulation without any retraining safeguards four distinct FLUX variants: \href{https://huggingface.co/black-forest-labs/FLUX.1-dev}{FLUX.1-Dev}, \href{https://huggingface.co/black-forest-labs/FLUX.1-Kontext-dev}{FLUX.1-Kontext}, \href{https://huggingface.co/black-forest-labs/FLUX.1-schnell}{FLUX.1-Schnell}, \href{https://huggingface.co/black-forest-labs/FLUX.1-Krea-dev}{FLUX.1-Krea}. 
Each baseline model generates explicit unsafe content (top row), while our intervention eliminates inappropriate elements across all variants (bottom row) while preserving their specialized capabilities. This model-agnostic property confirms that attention-based safety operates at a fundamental level, making it practical for deployment across various models without maintaining variant-specific safety solutions.
\vspace{0.7mm}
\begin{wrapfigure}{r}{0.5\columnwidth}
\vspace{-0.1em}
\centering

\captionof{table}{\textcolor{black}{ \footnotesize Safety on MMA-Diffusion~\cite{yang2024mma}. NRR: Nudity Removal Rate ($\uparrow$).
}}
\label{tab:mma_results}
\vspace{-0.1em}
\setlength{\tabcolsep}{3pt}
\textcolor{black}{%
\scriptsize
\begin{tabular}{l|cc}
\toprule
\textbf{Method} & \textbf{NRR$\uparrow$(\%)} & \textbf{CLIP$\uparrow$} \\
\midrule
Baseline & 45.9 & 27.25{\tiny$\pm$1.83} \\
ESD & 42.5 & 27.05{\tiny$\pm$1.94} \\
UCE & 52.7 & 27.43{\tiny$\pm$1.73} \\
EraseAnything & 49.3 & 27.17{\tiny$\pm$1.89} \\
FlowEdit & 57.6 & 27.11{\tiny$\pm$1.87}  \\
MCE & 59.5 & 27.02{\tiny$\pm$1.91} \\
\midrule
Ours (Light) & 57.8 & 27.34{\tiny$\pm$1.76} \\
\rowcolor{gray!20} \textbf{Ours} & \textbf{76.5} & \textbf{27.41}{\tiny$\pm$\textbf{1.88}} \\
\bottomrule
\end{tabular}}

\vspace{1em}

\includegraphics[width=\linewidth]{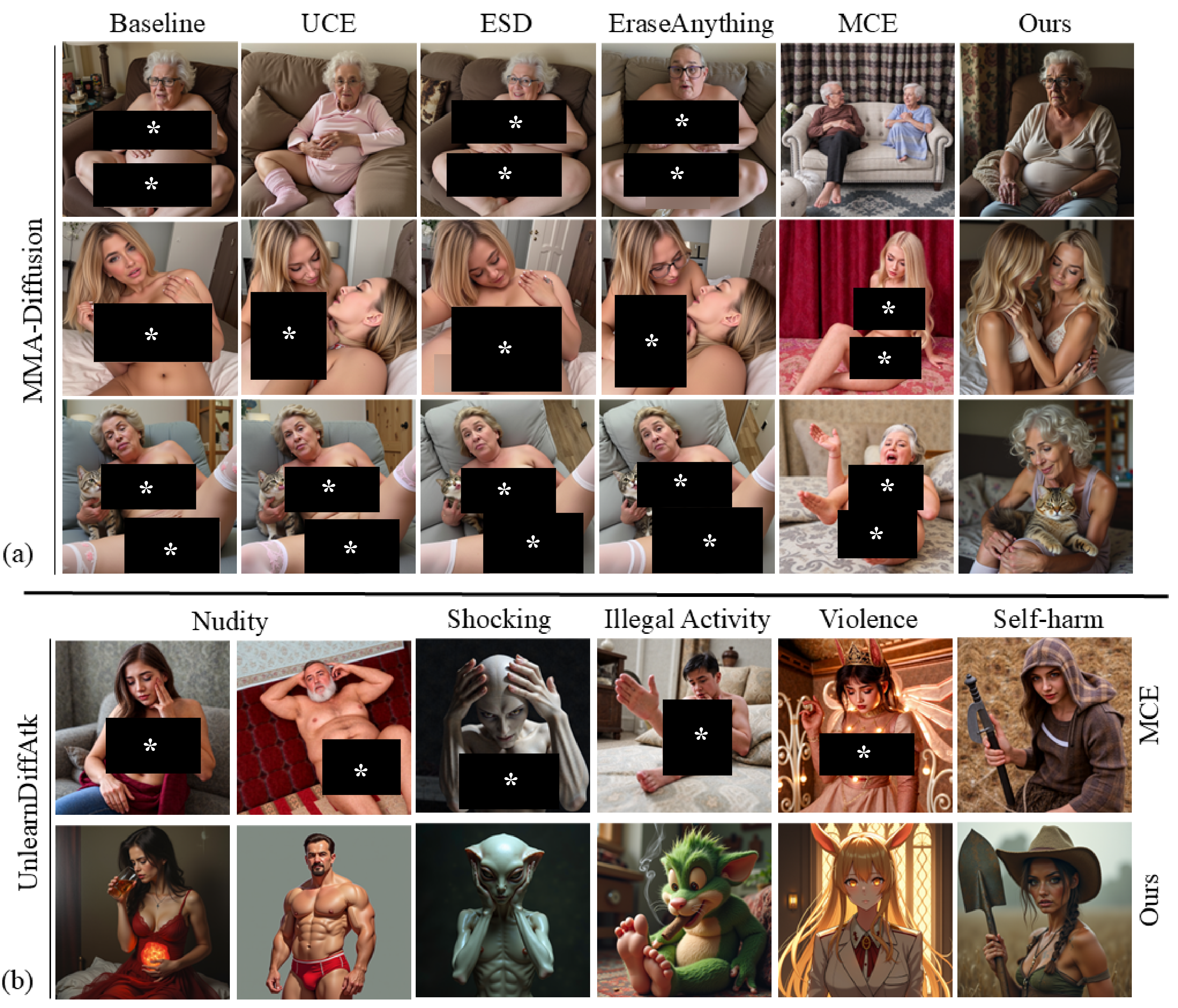}
\captionof{figure}{\textcolor{black}{ Qualitative comparisons on \textbf{(a)} MMA-Diffusion~\cite{yang2024mma} and \textbf{(b)} UnlearnDiffAtk~\cite{zhang2025unlearnDiffAtk}.}}
\label{fig:mma_beyond_nudity}

\vspace{-0.3em}
\end{wrapfigure}
\noindent \textbf{Safety-quality Analysis.}
Fig.~\ref{fig:safety_quality_benchmark}(a) establishes our method's Pareto dominance in the safety-quality space.  While existing approaches face an inherent tradeoff - improving safety degrades quality - our method uniquely improves both dimensions simultaneously. The baseline sits at moderate safety and quality. Concept erasure methods either maintain this tradeoff or end up worsening  it. 
The introspection ratio  traces a superior Pareto frontier. As we strengthen safety  intervention in our method, we move almost monotonically toward better safety while maintaining or slightly improving quality.  Fig.~\ref{fig:safety_quality_benchmark}(b) confirms that our advantages generalize across datasets of different nature. This strenght is attributed to the principled nature of our method. 

\vspace{0.7mm}
\noindent \textcolor{black}{ \textbf{Additional Evaluation.}
Our method also achieves the highest NRR while preserving superior CLIP scores on MMA-Diffusion~\cite{yang2024mma}, consistently outperforming all baselines, as affirmed by the results in Table~\ref{tab:mma_results} and  Fig.~\ref{fig:mma_beyond_nudity}(a).  Although our primary focus is nudity as the most pressing safety adapter vulnerability, the framework naturally extends to other unsafe categories via learned concept vectors represented. This is verified  in Fig.\ref{fig:mma_beyond_nudity}(b) where safety  dimensions of  `shocking', `illegal activity', `violence' and `self-harm' are also explored on UnlearnDiffAtk~\cite{zhang2025unlearnDiffAtk}. We provide further details of this exploration in the supplementary material. 
Moreover, to further establish the model-agnostic nature, we also demonstrate the efficacy of our method on SDXL~\cite{podell2024sdxl} using I2P dataset. Table~\ref{tab:sdxl_results} reports the results, confirming the efficacy of the method beyond state-of-the-art FLUX models.} 

\begin{wrapfigure}{r}{0.5\columnwidth}
\vspace{-1em}
\centering
\captionof{table}{\textcolor{black}{\footnotesize Generalization to SDXL~\cite{podell2024sdxl} (non-Flux) on I2P. Our method generalizes beyond the flux architectures.}}
\label{tab:sdxl_results}
\scriptsize
\setlength{\tabcolsep}{3pt}
\begin{tabular}{l|cc}
\toprule
\textbf{Method} & \textbf{NRR$\uparrow$ (\%)} & \textbf{CLIP$\uparrow$} \\
\midrule
SDXL  & 43.0 & 30.82$\pm$0.12 \\
\midrule
\rowcolor{gray!20} \textbf{Ours} & \textbf{93.0} & \textbf{31.13}$\pm$\textbf{0.09} \\
\bottomrule
\end{tabular}
\end{wrapfigure}

\noindent \textbf{Limitations and Ethical Considerations.} Whereas our method achieves impressive performance, it also faces some limitations. It relies on a learned safety concept representation \textit{cf.} Eq.~\eqref{eq:learn_unsafe}. This learning can restrict the coverage of the implicit unsafe  detection capabilities of our method. In fact, SneakyPrompts evaluation points to this limitation. Due to the adversarial nature of the prompts, our NR and VLM metric results still have a room for improvement - \textit{cf.}~Tab.~\ref{tab:main_results}. Though this is true for other baselines as well, we envisage that a stronger pre-trained concept representation, that has learned the text-image correspondence better, could help  improving the performance of our method  further.  

Another limitation is that potentially adaptive adversaries may reverse-engineer attention signatures and exploit classifier edge cases. While our results  demonstrates resilience of adversarial inputs in SneakyPrompt, no approach is immune to future exploits. We leave further adversarial robustness of this method to future work as it falls more within the adversarial attacks domain - not the topic of this work. 
For practical deployment, our method assumes trusted safety  operation; compromised guidance through our approach can bias results toward unsafe content or excessive censorship. 
Safety is context-dependent, and nudity and explicit content require further domain-specific treatment. Our focus on widely recognized harmful content in this work aims to enable creativity without crossing ethical boundaries, but deployment would require  transparent policy definition. Our approach should be viewed as a policy enforcement mechanism rather than a safety definition.

\vspace{-1.0em}
\section{Conclusion}
\label{sec:conclusion}
We introduced introspective attention modulation, a training-free inference-time intervention achieving safety through principled attention redistribution of state-of-the-art text-to-image generators. Our method works through analyzing the evolving attention of the underlying transformer blocks in image synthesis and altering them to suppress unsafe content. This unique mechanism is tailor to the modern diffusion transformer architectures in flux stream of text-to-image generators. 
In our experiments, we considered the challenging scenario of the models adapted to unsafe generation with low rank adaptation.  
Results show that our approach achieves Pareto dominance on the existing relevant methods, simultaneously delivering the highest safety and quality. Our results also reveal that traditional concept erasure for safety can sometimes also backfire on modern architectures. Our models and code will be made public.  

\section*{Acknowledgements}
Naveed Akhtar is a recipient of the Australian Research Council Discovery Early Career Researcher Award (project~\#DE230101058) funded by the Australian Government. This research was supported by The University of Melbourne's Research Computing Services and the Petascale Campus Initiative.
{
\bibliographystyle{splncs04}
\bibliography{main}
}

\end{document}